\documentclass[sigconf,nonacm]{acmart}

\usepackage{graphicx}
\usepackage{booktabs}
\usepackage{multirow}
\usepackage{float}
\usepackage{stfloats}

\AtBeginDocument{%
  }

\title{STREAM: A Data-Centric Framework for Mining High-Value Task-Oriented Dialogues from Streaming Media}

\author{Liang Xue}
\affiliation{%
  \institution{Harbin Institute of Technology}
  \department{Byering Technology}
  \city{Harbin}
  \country{China}}
\email{xueliang.xl@stu.hit.edu.cn}

\author{Haoyu Liu}
\affiliation{%
  \institution{Harbin Institute of Technology}
  \city{Harbin}
  \country{China}}

\author{Cheng Wang}
\affiliation{%
  \institution{Harbin Institute of Technology}
  \city{Harbin}
  \country{China}}

\author{Pengyu Chen}
\affiliation{%
  \institution{Harbin Institute of Technology}
  \city{Harbin}
  \country{China}}

\author{Haozhuo Zheng}
\affiliation{%
  \institution{Harbin Institute of Technology}
  \city{Harbin}
  \country{China}}

\author{Yang Liu}
\affiliation{%
  \institution{Harbin Institute of Technology}
  \city{Harbin}
  \country{China}}
\email{liuyang@hit.edu.cn}

\renewcommand{\shortauthors}{Xue et al.}

\begin{abstract}
Large language models for vertical domains are bottlenecked by the scarcity of complex, domain-specific task-oriented dialogues. Existing data acquisition pipelines face a persistent trilemma: expert annotation is expensive, real-world service conversations are constrained by privacy and commercial restrictions, and static corpora quickly become temporally stale. We propose Stream, a data-centric framework that leverages publicly available streaming media (live streams and short videos) to synthesize high-value service dialogues at scale. Stream mines authentic interaction signals from noisy streams and synthesizes conversations by integrating role-grounded persona construction with Conversational Blueprint construction; it further adopts retrieval-augmented generation (RAG) to support knowledge-aware responses. Based on Stream, we release StreamDial, a large-scale multi-domain dataset covering Automotive, Restaurant, and Hotel. StreamDial contains 87,498 dialogue sessions and 1,497,320 turns in total, with an average of 17.11 turns per session and a comparable scale across domains. Each session is organized as a structured quadruplet $\langle P_u, P_a, B, H \rangle$ that pairs dialogue history with explicit user/agent personas and a Conversational Blueprint, capturing realistic service behaviors such as requirement mining, constraint conflicts, negotiation, and recovery. Evaluations with automatic judges and downstream tasks show that StreamDial improves intrinsic dialogue quality over strong baselines, and models trained with StreamDial improve Dialogue State Tracking across backbones; we further report a completed human-evaluation set and encouraging multilingual transfer on Qwen3-8B under a controlled training budget. The data is released in https://github.com/hitxueliang/DialogDataSetBySTREAM.
\end{abstract}



\begin{document}
\maketitle

\begin{figure*}[t]
    \centering
    \includegraphics[width=\textwidth]{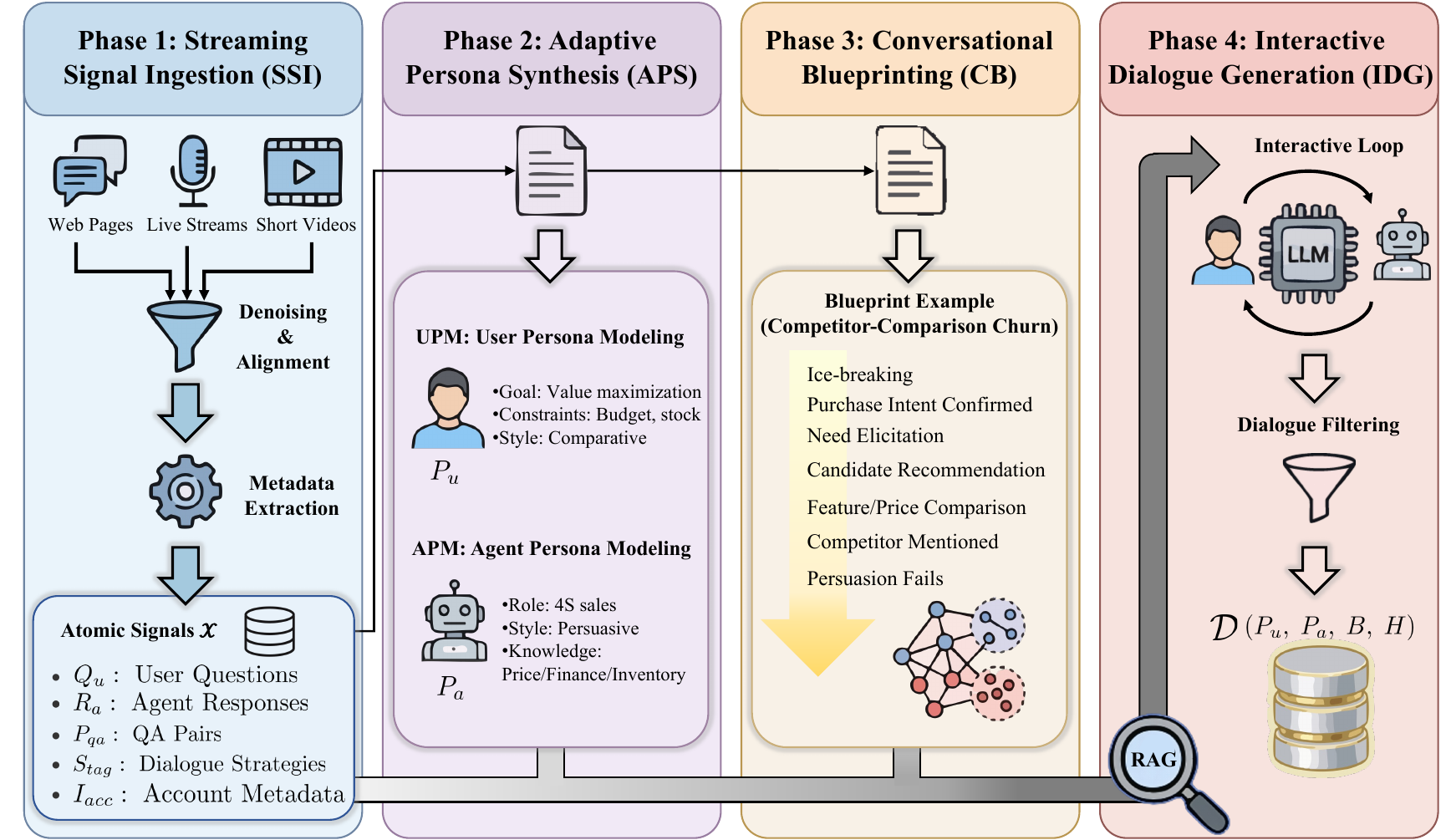}
    \caption{
    Overview of the Stream framework.
    Starting from heterogeneous rich-media sources (web pages, live streams, and short videos), 
    Phase 1 (\textbf{SSI}) extracts atomic interaction signals 
    $\mathcal{X}=\{Q_u, R_a, P_{qa}, S_{tag}, I_{acc}\}$ via denoising, alignment, and metadata extraction.
    Phase 2 (\textbf{APS}) builds paired role-grounded personas, including User Persona Modeling ($P_u$) and Agent Persona Modeling ($P_a$).
    Phase 3 (\textbf{CB}) constructs a Conversational Blueprint ($B$) that specifies stage-wise rhythm, key nodes, coping strategies, and feasible dialogue paths.
    Phase 4 (\textbf{IDG}) performs interactive multi-agent generation with RAG support and dialogue filtering, producing the final dataset
    $\mathcal{D}=\{(P_u, P_a, B, H)\}$.
    }
    \label{fig:pipeline}
\end{figure*}

\section{Introduction}
\label{sec:intro}

The rapid evolution of large language models has amplified a data scarcity paradox: the demand for high-quality, domain-specific training data continues to grow, while publicly available task-oriented dialogue corpora remain limited~\cite{budzianowski2018multiwoz,rastogi2020schema,zhu2020crosswoz,quan2020risawoz}. This challenge is especially pronounced in vertical service scenarios, where an assistant must go beyond answering questions and instead perform requirement mining, handle constraint conflicts, negotiate feasible options, and proactively guide users toward actionable outcomes~\cite{rastogi2020schema}. Such complex task-oriented dialogues require strategic coherence and professional knowledge that are difficult to obtain at scale.

However, collecting high-value vertical dialogues faces a persistent trilemma. Expert annotation is expensive and hard to scale~\cite{wen2017network}. Real-world service conversations are often confined to private channels and restricted by privacy and commercial constraints~\cite{Voigt2024TheEG}. Public static corpora also become temporally stale, making it difficult for models to reflect continuously evolving products, policies, and service practices~\cite{lewis2020rag,ram2023context}. As a result, models trained on existing datasets may perform well on canonical slot-filling setups yet struggle with realistic service behaviors such as inventory checking, reservation holding, or recovery after unmet constraints.

In this work, we explore publicly available streaming media as a scalable and timely data source. Live streams and short videos have become a major venue for professional service interactions, where experts respond to user questions in real time and comments reflect diverse intents and long-tail concerns. These interactions naturally expose service strategies, decision-making patterns, and domain knowledge that are difficult to capture through templated simulation or crowdsourcing alone~\cite{wang2022selfinstruct}.

We propose Stream, a data-centric framework that mines high-value interaction signals from streaming media and synthesizes complex task-oriented dialogues. Stream combines role-grounded persona construction with Conversational Blueprint construction and adopts retrieval-augmented generation (RAG) to support knowledge-aware responses during dialogue synthesis. Concretely, the framework performs a progressive transformation from noisy interaction signals to paired personas, then to blueprint-guided trajectories, and finally to executable multi-turn dialogues. Based on Stream, we release StreamDial, a large-scale multi-domain dataset covering Automotive, Restaurant, and Hotel. StreamDial contains 87,498 dialogue sessions and 1,497,320 turns in total, with comparable scale across domains. Importantly, each session is organized as a structured quadruplet $\langle P_u, P_a, B, H \rangle$, pairing dialogue history with explicit user and agent personas and a Conversational Blueprint, enabling supervision of role-conditioned and blueprint-guided service behaviors.

Our contributions are summarized as follows:
\begin{itemize}
    \item \textbf{A streaming-media-driven synthesis framework.} We introduce Stream, a general pipeline that transforms low-cost, unstructured streaming media into structured task-oriented dialogues by mining interaction signals and synthesizing conversations with role-grounded personas, Conversational Blueprints, and RAG-supported generation.
    \item \textbf{The StreamDial dataset.} We release StreamDial, a large-scale dataset spanning three vertical domains with \textbf{87,498} sessions and \textbf{1,497,320} turns. Each session follows a quadruplet structure $\langle P_u, P_a, B, H \rangle$, capturing complex service behaviors including requirement mining, constraint conflicts, negotiation, and recovery.
    \item \textbf{Comprehensive intrinsic and extrinsic evaluation.} We evaluate dialogue quality with automatic LLM-based assessment, report an independent human-validation protocol with a completed human-evaluation set, and validate downstream utility on Dialogue State Tracking using Joint Goal Accuracy and Slot-value F1. Results show consistent gains over strong baselines and encouraging multilingual transfer under a controlled training budget.
\end{itemize}

\section{Related Work}
\label{sec:related_work}

\noindent\textbf{Task-oriented Dialogue Datasets}\quad 
The advancement of TOD systems is fundamentally driven by the quality and availability of annotated corpora. Seminal benchmarks like MultiWOZ \cite{budzianowski2018multiwoz} and SGD \cite{rastogi2020schema} laid the groundwork for multi-domain state tracking, while subsequent datasets such as CrossWOZ \cite{zhu2020crosswoz} and TransferTOD \cite{zhang2024transfertod} introduced deeper complexity through cross-domain dependencies. However, relying on crowdsourced construction creates inherent bottlenecks \cite{peng2020soloist,zhang2020todbert}. First, the prohibitive cost of manual annotation limits the scale of these datasets \cite{rastogi2020schema}. Second, crowd workers often lack the domain expertise required to simulate professional consultants, resulting in interactions that are functionally correct but strategically shallow—missing key elements like persuasion, up-selling, and objection handling. Furthermore, these static benchmarks suffer from severe temporal rigidity; they fail to capture dynamic domain shifts, such as evolving vehicle models or updated legal regulations, rendering them less effective for training agents meant for real-time deployment. StreamDial overcomes these limitations by tapping into the live ecosystem of streaming media, ensuring both strategic depth and temporal freshness.

\vspace{0.5em}

\noindent\textbf{Persona-Driven Simulation and Synthesis}\quad 
To mitigate data scarcity, synthetic generation methods have increasingly shifted toward persona-driven simulation. Recent works such as Generative Agents \cite{park2023generative} and PatientSim \cite{kyung2025patientsim} attempt to improve realism by defining specific cognitive profiles. However, these simulations often suffer from interaction asymmetry and strategy-action decoupling. First, user personas in these frameworks are typically sampled from predefined templates or the internal knowledge of LLMs, failing to capture the dynamic intent shifts and unpolished linguistic traits of real-world vertical users. Second, the absence of explicit professional constraints leads to generic agent behaviors that lack the specific service strategies required for high-stakes domains like automotive sales or legal consulting. Most importantly, existing methods often treat dialogue as a random-walk process rather than a goal-oriented progression, resulting in conversations that lack a clear roadmap. Stream addresses these limitations by decomposing dialogue synthesis into three grounded components: User Persona Modeling for capturing authentic intent distributions, Agent Persona Modeling for defining professional service boundaries, and Conversational Blueprinting (CB) to ensure the strategic coherence of the interaction. By grounding these modules in real-world streaming signals, we move beyond simplistic profile-based generation toward structured, logic-driven simulation.

\vspace{0.5em}
\noindent\textbf{Mining Knowledge from Rich Media}\quad 
The extraction of value from unstructured rich media is an expanding frontier \cite{radford2021learning,alayrac2022flamingo}. Pipelines such as EuroSpeech \cite{pfisterer2025eurospeech} and Data-Juicer \cite{chen2025datajuicer} have demonstrated the feasibility of aligning multimodal signals and cleaning massive datasets at scale. However, prior efforts have predominantly focused on modality alignment (e.g., ASR, translation) rather than semantic reconstruction \cite{radford2023whisper,ko2023seamlessm4t}. To date, streaming media remains an underutilized resource for extracting high-level dialogue logic. Stream treats livestreams not merely as audio signals, but as repositories of Atomic Interaction Signals. By systematically mining expert strategies and user intent shifts from these streams, we bridge the gap between noisy raw media and structured, logic-rich dialogue data suitable for training sophisticated TOD agents.

\section{The Stream Framework}
\label{sec:stream_framework}

We propose Stream, a data-centric framework designed to mine high-value, domain-specific conversations from unstructured rich media. 
Formally, let $\mathcal{S} = \{s_1, s_2, \dots, s_M\}$ denote a continuous stream of heterogeneous input data. 
Our objective is to synthesize a structured dialogue dataset 
$\mathcal{D} = \{(P_u^{(i)}, P_a^{(i)}, B^{(i)}, H^{(i)})\}_{i=1}^N$, 
where each sample $d^{(i)} \in \mathcal{D}$ is a quadruplet 
$\langle P_u^{(i)}, P_a^{(i)}, B^{(i)}, H^{(i)} \rangle$.
Here, $P_u$ denotes the user persona, $P_a$ the agent persona, $B$ the Conversational Blueprint, and $H$ the synthesized dialogue history.

As illustrated in Figure~\ref{fig:pipeline}, the framework consists of four cascaded phases: 
Streaming Signal Ingestion (SSI), Adaptive Persona Synthesis (APS), Conversational Blueprinting (CB), and Interactive Dialogue Generation (IDG).

\subsection{Phase 1: Streaming Signal Ingestion (SSI)}
SSI serves as the foundational ingestion engine that harvests atomic interaction signals from unstructured rich media \cite{chen2025datajuicer,pfisterer2025eurospeech}. 
Its input is $\mathcal{S}$, which aggregates three heterogeneous sources: 
Web Pages containing static domain knowledge; 
Live Streams featuring real-time audio-visual content and synchronized bullet chats; 
and Short Videos including edited clips and comment sections. 
Unlike static corpora, these sources contain large-scale real-time interactions (e.g., bullet chats and live calls), enabling the framework to capture dynamic interplay between users and hosts.

From these raw inputs, SSI extracts five types of atomic interaction signals, denoted as $\mathcal{X}$:
\begin{itemize}
    \item User Questions ($Q_u$): isolated by denoising bullet chats and comments to retain entries with clear and authentic intent.
    \item Agent Responses ($R_a$): extracted from transcribed host speech via automated speech recognition \cite{radford2023whisper}, or retrieved from textual replies.
    \item QA Pairs ($P_{qa}$): formed by aligning $Q_u$ and $R_a$ using temporal proximity and semantic relevance \cite{reimers2019sentencebert}, ensuring logical consistency.
    \item Dialogue Strategies ($S_{tag}$): identified by a strategy classifier to capture response and guidance logic in host--user interactions.
    \item Account Metadata ($I_{acc}$): extracted from profile and certification information to anchor professional context and service boundaries.
\end{itemize}

\vspace{0.4em}
\noindent\textbf{Signal Cleaning and ASR Control}\quad
Because streaming media is noisy, SSI applies a multi-stage quality-control procedure before downstream synthesis.
For speech content, we use a commercial ASR system and observe a word error rate ranging from 3.5\% to 10.5\% across sampled source materials.
To reduce domain-specific transcription errors, the ASR output is normalized with a domain lexicon, including a 4.1M-entry automotive vocabulary and 43K address entries for location-sensitive hotel and restaurant content.
We then apply retrieval-based evidence checking, LLM-assisted correction, and consistency validation to remove or repair mismatched entity names, prices, dates, locations, and configuration terms.
Finally, candidate QA pairs are retained only when temporal proximity, semantic relevance, and domain-entity consistency are jointly satisfied.
This procedure is designed to prevent ASR and comment-noise artifacts from being directly propagated into personas, blueprints, or generated dialogues.

\subsection{Phase 2: Adaptive Persona Synthesis (APS)}
To bridge the gap between abstract simulation and realistic interaction, we propose Adaptive Persona Synthesis (APS) to construct high-fidelity representations for both dialogue participants \cite{park2023generative,kyung2025patientsim}. 
The output of this phase is a paired persona set $(P_u, P_a)$, which provides role constraints and behavioral priors for downstream blueprint construction and dialogue generation.

\vspace{0.5em}

\noindent \textbf{User Persona Modeling (UPM)}\quad 
UPM synthesizes a structured user representation $P_u$ by integrating real-time user questions $Q_u$ with representative seed dialogues \cite{zhu2020crosswoz,zhang2024transfertod}. 
Each $P_u$ is defined by a clear objective, encompassing basic information (e.g., power type, budget), core requirements (e.g., seat heating), and primary inquiries. 
To facilitate authentic simulation, UPM also generates potential utterances that reflect natural language variations of user intent, ensuring that the simulated agent encounters realistic linguistic diversity \cite{wang2022selfinstruct}. 
In practice, this design helps preserve both intent-level consistency (what the user wants) and surface-level variability (how the user expresses it).

\vspace{0.5em}

\noindent \textbf{Agent Persona Modeling (APM)}\quad 
Complementarily, APM defines the agent representation $P_a$ by leveraging account metadata $I_{acc}$ and professional interaction patterns \cite{park2023generative,kyung2025patientsim}. 
This modeling process captures identity positioning, linguistic style, and service boundaries. 
A cornerstone of $P_a$ is the integration of a domain-specific knowledge base $\mathcal{R}$, which provides technical specifications for candidate options (e.g., engine performance and feature availability). 
This paired architecture between $P_u$ and $P_a$ establishes a goal-oriented interaction environment in which the agent's expertise is aligned with the user's structured constraints. 
As a result, APS provides stable role grounding for subsequent Conversational Blueprint construction.

\subsection{Phase 3: Conversational Blueprinting (CB)}
To maintain logical consistency in long-context interactions, CB constructs a Conversational Blueprint $B$ before dialogue synthesis begins \cite{young2013pomdp,williams2017hybrid}. 
$B$ serves as a strategic and executable interaction specification, rather than a simple turn sequence \cite{yao2023react,yao2023tree}. 
This phase transforms the extracted signals $\mathcal{X}$ (particularly strategy signals $S_{tag}$), together with agent persona $P_a$ and multiple seed dialogues, into a coherent blueprint for goal-oriented interaction \cite{budzianowski2018multiwoz,rastogi2020schema}.

As illustrated in our framework, the generated Conversational Blueprint $B$ consists of four hierarchical components:
\begin{itemize}
    \item Overall Rhythm Overview: Defines the progressive stages of the dialogue, such as intent identification and requirement mining, ensuring a clear macro-level trajectory.
    \item Key Node Definitions: Specifies the business meaning and trigger value of user signals, guiding transitions between conversational states.
    \item Typical Scenarios and Coping Strategies: Outlines specific conversational actions and linguistic tactics for handling diverse situations such as brand comparisons.
    \item Dialogue Flow Atlas: Maps multiple interaction paths, providing a graph-structured view of feasible conversational outcomes.
\end{itemize}

This blueprinting step provides explicit trajectory-level guidance for the subsequent generation phase, reducing random-walk behavior in long dialogues and improving strategic consistency across turns.

\subsection{Phase 4: Interactive Dialogue Generation (IDG)}
This final phase serves as the core synthesis engine, transforming $P_u$, $P_a$, and $B$ into high-fidelity, goal-oriented interactions through multi-agent simulation and graph-based refinement \cite{park2023generative,wang2024agents}. 
The output is the synthesized dialogue history $H$, with each turn grounded in persona constraints and the Conversational Blueprint.

\noindent\textbf{RAG-Enhanced Dialogue Synthesis}\quad 
The workflow begins with the \textbf{Dialogue Configuration Selector}, which samples a $P_u$ and retrieves the most compatible $P_a$ based on a matching score. 
Subsequently, \textbf{Dialogue Opening Synthesis (DOS)} initiates the exchange by using \textbf{RAG} to extract and rewrite opening patterns from seed dialogues \cite{lewis2020rag,izacard2022few}. 
To ensure realistic progression, we implement a bidirectional interactive loop:
\begin{itemize}
    \item \textbf{User Utterance Simulation (UUS)}: Based on history $h_{t-1}$, UUS employs RAG to retrieve agent messages similar to $a_{t-1}$ from the retrieval pool \cite{lewis2020rag,asai2024selfrag}. The real user responses following these messages serve as behavioral reference samples $\mathcal{E}_u$. UUS generates a simulated message $u_t$ alongside a structured inform block to capture user constraints.
    \item \textbf{Agent Response Generation (ARG)}: Symmetrically, ARG utilizes RAG to retrieve user queries similar to $u_t$. The corresponding expert responses act as professional evidence $\mathcal{E}_a$. By referencing these samples, ARG produces a response $a_t$ that incorporates domain knowledge $\mathcal{R}$ and a structured request block, driving the conversation forward according to $B$ \cite{ram2023context}.
\end{itemize}

\vspace{0.5em}

\noindent\textbf{Graph-Based Dialogue Filtering (DF)}\quad 
To ensure dataset quality and diversity, we construct a similarity graph where each node is a complete dialogue $D$. 
An edge exists between $D_i$ and $D_j$ only if both their aggregated user-side representation $\mathbf{u}$ and agent-side representation $\mathbf{a}$ exceed semantic thresholds:
\begin{equation}
    \text{Edge}(D_i, D_j) = \mathbb{I}(\text{sim}(\mathbf{u}_i, \mathbf{u}_j) > \tau_u) \wedge \mathbb{I}(\text{sim}(\mathbf{a}_i, \mathbf{a}_j) > \tau_a)
\end{equation}
By performing community detection on this graph, we identify clusters of redundant interactions and sample proportionally from each cluster to reduce redundancy while preserving nuanced diversity.

\begin{figure*}[htb]
    \centering
    \small
    \setlength{\tabcolsep}{6pt}
    \renewcommand{\arraystretch}{1.15}

    \begin{tabular}{p{0.48\textwidth} p{0.48\textwidth}}
    \toprule
    \textbf{User Persona $P_u$ (excerpt)} & \textbf{Agent Persona $P_a$ (excerpt)} \\
    \midrule
    \textbf{Mindset:} rational and price-sensitive; frequently compares model years and promotions. &
    \textbf{Role:} brand consultant; provides model details, inventory checks, and conversion guidance. \\
    \textbf{Key concerns:} inventory availability, production date, total cost breakdown, and model-year differences. &
    \textbf{Style:} professional yet friendly; uses light persuasion and urgency cues when appropriate. \\
    \textbf{Primary goal:} decide whether the 2025 clearance version is worth buying versus waiting for a potential 2026 update. &
    \textbf{Service scope:} inventory lookup, configuration explanation, and test-drive invitation. \\
    \bottomrule
    \end{tabular}

    \vspace{0.6em}

    \begin{tabular}{p{0.96\textwidth}}
    \toprule
    \textbf{Dialogue $H$ (excerpt; Automotive)} \\
    \midrule
    \textbf{U:} Is the 2025 clearance Q5L worth buying? Is there a major tech gap versus the 2026 model? \\
    \textbf{A:} The 2025 clearance version is essentially the current main lineup. The 2026 model has not been officially released yet and is unlikely to introduce a dramatic change immediately, while the clearance price is typically more cost-effective. \\
    \textbf{U:} How many units are in stock? Can you check the production date? \\
    \textbf{A:} We have around three units in stock, produced in late 2024. I can check the exact details for you. \\
    \textbf{U:} Are all three units AWD? \\
    \textbf{A:} Yes, all three are quattro ultra AWD. \\
    \textbf{U:} How is the fuel consumption for the AWD version? \\
    \textbf{A:} In daily driving, it is usually around \emph{high single digits} in the city and \emph{mid-to-high single digits} on highways, depending on your route and driving style. What is your typical driving condition? \\
    \textbf{U:} That sounds a bit high for city driving. Do you have a 2WD option? \\
    \textbf{A:} For the clearance batch, we currently only have AWD. If fuel economy is your top priority, we can compare a more efficient 2WD alternative and arrange a test drive. \\
    \textbf{U:} Can I test drive it? \\
    \textbf{A:} Absolutely. What day works for you? I will reserve a test-drive vehicle. \\
    \bottomrule
    \end{tabular}

    \caption{
    Case study from StreamDial (Automotive). The excerpt shows realistic service operations (inventory and production-date inquiry), a mid-dialogue constraint conflict (fuel consumption vs.\ drivetrain availability), and transition to a feasible next step (alternative recommendation and test-drive invitation).
    The original dialogues are in Chinese; the excerpt shown here is translated for readability. The full Chinese version is provided in Appendix~\ref{app:case_study_zh}.
    }
    \label{fig:case_study_streamdial}
\end{figure*}

\section{The StreamDial Dataset}
\label{sec:streamdial_dataset}

StreamDial is a large-scale, multi-domain task-oriented dialogue dataset synthesized from publicly available streaming media.
It targets complex service scenarios where assistants must not only answer questions, but also conduct requirement mining, handle constraint conflicts, and proactively guide users toward actionable outcomes (e.g., test-drive booking, reservation, or check-in).
StreamDial covers three representative vertical domains: Automotive, Restaurant, and Hotel, reflecting diverse service patterns and decision-making processes.

\vspace{0.4em}
\noindent\textbf{Data collection and source filtering.}\quad
StreamDial is constructed by running the complete four-stage Stream pipeline on raw public rich-media sources, rather than on an already preprocessed dialogue corpus.
The monitored source pool contains more than 320K candidate public accounts, livestream rooms, short videos, and service pages from Douyin, Kuaishou, Xiaohongshu, and Ctrip.
We selected these platforms because they contain dense vertical-service interactions in the three target domains: vehicle consultation and sales, restaurant discovery and reservation, and hotel search and booking.

Source candidates are filtered with explicit domain and interaction criteria.
First, we keep accounts or content streams that match industry-specific keywords, category tags, account certifications, and service metadata.
Second, we prioritize high-interaction sources, including livestream rooms with more than 1,000 viewers or comparable comment activity, because such sources expose richer long-tail user intents.
Third, we remove low-signal content such as advertising-only clips, generic entertainment streams, repeated comments, and conversations without actionable service goals.
The retained sources are then passed to SSI for denoising, ASR correction, temporal alignment, and semantic consistency checks.
This filtering pipeline reduces source noise and makes the corpus more reproducible by separating source identification, quality filtering, and dialogue synthesis.

\begin{table}[htb]
    \centering
    \small
    \caption{Intermediate construction statistics used by StreamDial.}
    \label{tab:construction_stats}
    \setlength{\tabcolsep}{6pt}
    \begin{tabular}{lccc}
        \toprule
        \textbf{Domain} & \textbf{Agent Personas} & \textbf{User Personas} & \textbf{Blueprints} \\
        \midrule
        Automotive & 443 & 619 & 127 \\
        Restaurant & 229 & 4,420 & 150 \\
        Hotel & 227 & 4,446 & 148 \\
        \bottomrule
    \end{tabular}
\end{table}

Table~\ref{tab:construction_stats} reports the intermediate persona and blueprint scale after source filtering and signal extraction.
These artifacts are not latent generation prompts only; they are retained as structured fields associated with the released sessions, enabling inspection of role grounding and trajectory planning.

\vspace{0.4em}
\noindent\textbf{Dataset scale.}\quad
StreamDial contains 87,498 dialogue sessions with 1,497,320 turns in total, yielding 17.11 turns per session on average.
The three domains are constructed with comparable scale to support balanced evaluation and cross-domain analysis.
Table~\ref{tab:streamdial_stats} summarizes key dataset statistics.

\begin{table}[htb]
    \centering
    \small
    \caption{StreamDial dataset statistics. Numbers are reported at the session/turn level.}
    \label{tab:streamdial_stats}
    \setlength{\tabcolsep}{6pt}
    \begin{tabular}{lccc}
        \toprule
        \textbf{Domain} & \textbf{Dialogues} & \textbf{Turns} & \textbf{Avg. Turns / Dialogue} \\
        \midrule
        Automotive  & 29,486 & 566,095 & 19.20 \\
        Restaurant  & 27,389 & 450,703 & 16.46 \\
        Hotel       & 30,623 & 480,522 & 15.69 \\
        \midrule
        \textbf{Total} & \textbf{87,498} & \textbf{1,497,320} & \textbf{17.11} \\
        \bottomrule
    \end{tabular}
\end{table}

\vspace{0.4em}
\noindent\textbf{Session schema and fields.}\quad
Each StreamDial session is stored as a structured quadruplet $\langle P_u, P_a, B, H \rangle$, where
$P_u$ is the user persona, $P_a$ is the agent persona, $B$ is the dialogue plan, and $H$ is the multi-turn dialogue history.
This schema provides aligned supervision signals beyond plain text and supports analysis at both the role level and the trajectory level.

In the released data, persona fields are explicitly textual rather than latent tags.
For example, $P_u$ typically contains user mindset, core objectives, and major doubts, while $P_a$ records service role, communication style, business scope, and domain knowledge.
The plan field $B$ records stage-wise flow, key user signals, and candidate outcomes, enabling trajectory-level inspection of conversations.
The dialogue field $H$ stores turn-ordered utterances for model training and evaluation.

\vspace{0.4em}
\noindent\textbf{Observed interaction characteristics.}\quad
Compared with conventional public TOD corpora, StreamDial sessions more frequently include service-facing operations such as inventory checks, production-date inquiries, reservation/holding requests, freebie negotiation, and follow-up arrangement.
Many dialogues also show non-linear progress, where users revise requirements mid-conversation and the interaction transitions from one candidate solution to another.
This pattern is common in high-stakes service domains and provides realistic supervision for decision-oriented dialogue modeling.

\vspace{0.4em}
\noindent\textbf{Case study.}\quad
Figure~\ref{fig:case_study_streamdial} shows a representative Automotive example.
The excerpt illustrates a typical trajectory: initial model-year comparison, inventory and production-date confirmation, emergence of a constraint conflict (fuel consumption vs.\ drivetrain availability), and transition to an alternative option with test-drive scheduling.
The original dialogue is in Chinese; an English translation is shown in the main text, and the Chinese version is provided in Appendix~\ref{app:case_study_zh}.

\vspace{0.4em}
\noindent\textbf{Release and reproducibility.}\quad
The Chinese version of StreamDial is publicly released with the session-level quadruplet schema.
To support reproduction of the multilingual experiments in Section~\ref{sec:cross_lingual}, we are preparing translated English, French, and Korean versions using the same translation and post-processing pipeline as our experiments.
The English release is prioritized first, with automatic translation followed by 1\% manual sampling inspection for terminology, slot preservation, and dialogue-role consistency.
We will release the translated data together with translation prompts, post-processing rules, and split files, so that future work can reproduce the cross-lingual transfer setting without rebuilding the translation pipeline.

\section{Experiments}
\label{sec:experiments}

In this section, we evaluate StreamDial from three perspectives.
First, we assess the intrinsic quality of synthesized dialogues with an LLM-as-a-Judge protocol and complement it with human validation, using automatic judging as a scalable screening tool rather than a replacement for human assessment \cite{zheng2023judging,Gu2024ASO}.
Second, we evaluate extrinsic utility on downstream Dialogue State Tracking (DST), including both overall results and cross-domain generalization \cite{rastogi2020schema}.
Finally, we study multilingual transfer on X-RiSAWOZ to test whether the gains transfer across target languages \cite{Moradshahi2023XRiSAWOZHE}.

\subsection{Experimental Setup}
\label{sec:exp_setup}

\noindent\textbf{Public Open Data pool.}\quad
We construct an Open Data pool by mixing three publicly available Chinese task-oriented dialogue datasets:
CrossWOZ~\cite{zhu2020crosswoz}, RiSAWOZ~\cite{quan2020risawoz}, and TransferTOD~\cite{zhang2024transfertod}.
Unless otherwise specified, Open Data (Baseline) refers to human-annotated dialogues sampled from this pool (matched by domain when applicable).
For multilingual evaluation, we additionally use X-RiSAWOZ (few-shot/train split, validation, and test) in the Automotive domain, following its official split protocol \cite{Moradshahi2023XRiSAWOZHE}.
For cross-lingual experiments, the synthesized StreamDial samples are translated from Chinese into each target language (English, French, Korean) before training, so each run uses target-language public dialogues mixed with target-language translated synthetic dialogues under the same 2k budget.

\vspace{0.5em}
\noindent\textbf{Data configurations.}\quad
We compare four data configurations:
(1) Open Data (Baseline), the original public human-annotated dialogues from the Open Data pool;
(2) Seed-Only, dialogues synthesized from seed dialogues/templates without streaming signals, where seed indicates dialogues sampled from RiSAWOZ;
(3) Stream-Only, dialogues synthesized from stream signals in a zero-shot manner, where stream indicates interaction signals extracted from publicly available streaming media (e.g., live-stream rooms and short videos); and
(4) \textbf{StreamDial (Hybrid)}, our full pipeline that fuses seed scaffolding with stream signals.

\vspace{0.5em}
\noindent\textbf{Evaluation criteria.}\quad
We conduct both intrinsic and extrinsic evaluations:
\begin{itemize}
    \item \textbf{Intrinsic (Dialogue Quality).} We adopt an LLM-as-a-Judge protocol and score dialogues on a 1--10 scale along six dimensions:
    \textit{Coherence}, \textit{Informativeness}, \textit{Naturalness}, \textit{Diversity}, \textit{Flexibility}, and \textit{Overall Quality}.
    For automatic intrinsic scoring, we sample 2,000 dialogues per configuration, evenly distributed across domains, and use fixed prompts and decoding settings across judges for consistency.
    To reduce presentation bias, samples are anonymized, source labels are hidden, and candidate outputs are scored in randomized order under the same formatting template.
    We treat these scores as large-scale diagnostic signals and complement them with an independent human-evaluation protocol.
    \item \textbf{Extrinsic (DST Utility).} We evaluate downstream Dialogue State Tracking (DST) and report Joint Goal Accuracy (JGA) and Slot-value F1.
\end{itemize}

\vspace{0.5em}
\noindent\textbf{Judge models and DST backbones.}\quad
For intrinsic evaluation, we use three judge models: Qwen3-Max~\cite{qwen3_modelcard}, GPT-5.2~\cite{Achiam2023GPT4TR}, and Gemini3-Pro~\cite{Comanici2025Gemini2P}.
For downstream DST, we fine-tune three backbones---Qwen3-1.7B~\cite{qwen3_modelcard}, Qwen3-8B~\cite{qwen3_modelcard}, and Gemma3-4B~\cite{Kamath2025Gemma3T}---under the same training protocol to ensure fair comparison.

\begin{table}[h]
    \centering
    \small
    \caption{Cross-domain quality comparison (\textit{Overall Quality}) across Automotive, Restaurant, and Hotel domains.}
    \label{tab:domain_comparison}
    \setlength{\tabcolsep}{6pt}
    \begin{tabular}{l|cc|cc|cc}
        \toprule
         & \multicolumn{2}{c|}{\textbf{Qwen3-Max}} & \multicolumn{2}{c|}{\textbf{GPT-5.2}} & \multicolumn{2}{c}{\textbf{Gemini3-Pro}} \\
        \textbf{Domain} & Base & Ours & Base & Ours & Base & Ours \\
        \midrule
        Automotive & 6.32 & \textbf{8.98} & 5.89 & \textbf{7.81} & 6.91 & \textbf{8.82} \\
        Restaurant & 7.15 & \textbf{8.81} & 7.01 & \textbf{7.93} & 7.99 & \textbf{8.85} \\
        Hotel      & 7.31 & \textbf{8.72} & 7.03 & \textbf{7.89} & 7.96 & \textbf{8.92} \\
        \bottomrule
    \end{tabular}
\end{table}

\begin{table*}[h]
    \centering
    \small
    \caption{DST performance comparison under different training strategies (Public test set).}
    \label{tab:dst}
    \setlength{\tabcolsep}{7pt}
    \renewcommand{\arraystretch}{1.15}
    \begin{tabular}{l|cc|cc|cc}
        \toprule
        \textbf{Training Strategy}
        & \multicolumn{2}{c|}{\textbf{Qwen3-1.7B}}
        & \multicolumn{2}{c|}{\textbf{Qwen3-8B}}
        & \multicolumn{2}{c}{\textbf{Gemma3-4B}} \\
        & \textbf{JGA(\%)} & \textbf{F1(\%)}
        & \textbf{JGA(\%)} & \textbf{F1(\%)}
        & \textbf{JGA(\%)} & \textbf{F1(\%)} \\
        \midrule
        Zero-shot (No SFT) & 45.71 & 74.61 & 54.04 & 73.40 & 33.59 & 67.93 \\
        \midrule
        SFT w/ Public Data (Baseline) & 93.43 & 98.81 & 93.94 & 98.56 & 95.71 & 98.78 \\
        \midrule
        \hspace{0.8em}+ Seed-Only & 95.45 & 98.82 & 95.71 & 99.20 & 97.73 & 99.49 \\
        \hspace{0.8em}+ Stream-Only & 96.46 & 99.32 & 95.96 & 99.12 & 97.98 & 99.49 \\
        \hspace{0.8em}+ StreamDial (Hybrid) & \textbf{96.97} & \textbf{99.37} & \textbf{96.72} & \textbf{99.41} & \textbf{97.98} & \textbf{99.54} \\
        \bottomrule
    \end{tabular}
\end{table*}
\subsection{Intrinsic Evaluation: Dialogue Quality}
\label{sec:intrinsic_eval}

\begin{figure*}[htb]
    \centering
    \includegraphics[width=\textwidth]{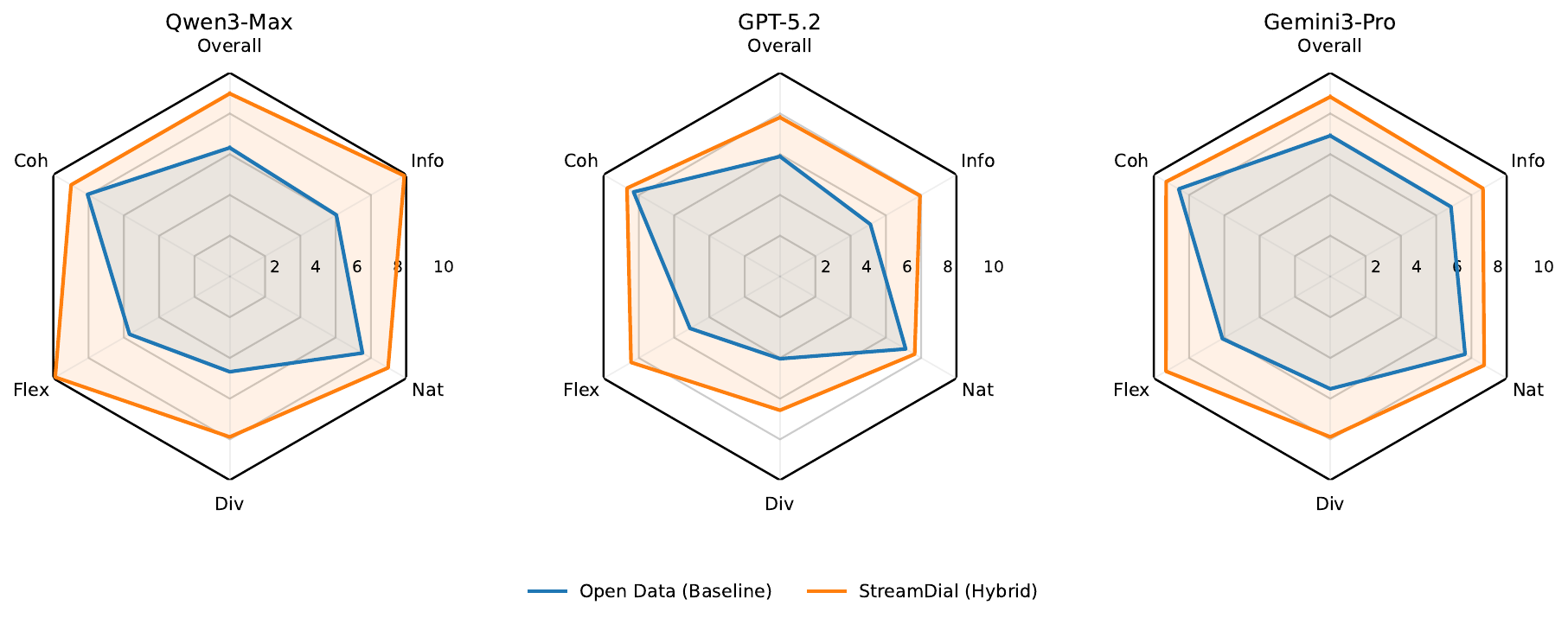}
    \caption{Radar visualization of intrinsic quality scores comparing Open Data (Baseline) and \textbf{StreamDial (Hybrid)} under three LLM judges (Qwen3-Max, GPT-5.2, Gemini3-Pro). Each axis corresponds to one evaluation dimension. The exact numeric values used to plot this figure are reported in Appendix~\ref{app:radar_scores}.}
    \label{fig:intrinsic_radar}
\end{figure*}

We first evaluate intrinsic dialogue quality with an LLM-as-a-Judge protocol.
Each dialogue is scored on a 1--10 scale over six dimensions: Coherence, Informativeness, Naturalness, Diversity, Flexibility, and Overall Quality.
Figure~\ref{fig:intrinsic_radar} compares Open Data (Baseline) and StreamDial (Hybrid) under three independent judges (Qwen3-Max, GPT-5.2, and Gemini3-Pro).

Across all three judges, StreamDial (Hybrid) shows a clear advantage in overall quality.
The gains are especially visible on Informativeness, Diversity, and Flexibility, indicating that stream-grounded synthesis produces conversations with richer information, broader interaction patterns, and more adaptive turns.
At the same time, Coherence and Naturalness remain strong, suggesting that the additional strategic complexity does not hurt conversational fluency or logical continuity.

An important observation is that the trend is stable across different judge families.
Although absolute scores differ across evaluators, the relative ordering between baseline and StreamDial is consistent, which reduces the risk that the improvements are artifacts of a single judge model.
To support reproducibility, we provide the full per-dimension score table used for Figure~\ref{fig:intrinsic_radar} in Appendix~\ref{app:radar_scores}.

We further test whether these improvements hold across domains.
Table~\ref{tab:domain_comparison} reports Overall Quality on Automotive, Restaurant, and Hotel.
StreamDial (Hybrid) improves domain-level scores in all three domains under all judges, showing that the intrinsic quality gains are not tied to a single service scenario.
This cross-domain consistency aligns with the design of our pipeline: personas are grounded in real interaction signals, and dialogue trajectories are guided by explicit blueprints rather than shallow template expansion.

\vspace{0.5em}
\noindent\textbf{Human validation.}\quad

To address the limitations of automatic judging, we conduct an independent human evaluation of dialogue quality.
The evaluated set contains 1,431 completed sessions sampled from the three domains and uses the same six dimensions as the automatic protocol: Coherence, Informativeness, Naturalness, Diversity, Flexibility, and Overall Quality.
Sessions are anonymized and presented in randomized order without source labels, so annotators do not know whether a session comes from public data or StreamDial.
Annotators are asked to inspect the full session, including user persona, agent persona, blueprint, and dialogue history, and then score each dimension on a 1--10 scale.
They are also instructed to flag factual inconsistency, role violation, unnatural language, and unsupported service actions.

\begin{table}[h]
    \centering
    \small
    \caption{Human evaluation protocol and completed evaluation set.}
    \label{tab:human_eval_protocol}
    \setlength{\tabcolsep}{6pt}
    \begin{tabular}{lc}
        \toprule
        \textbf{Item} & \textbf{Value} \\
        \midrule
        Evaluated sessions & 1,431 \\
        Annotators involved & 53 \\
        Evaluation dimensions & 6 \\
        Presentation & Anonymized / randomized \\
        Domains & Automotive / Restaurant / Hotel \\
        \bottomrule
    \end{tabular}
\end{table}

Table~\ref{tab:human_eval_protocol} summarizes the protocol and the completed annotation set used in this version.
This set contains 1,431 sessions annotated by 53 annotators.
The human results are consistent with the automatic evaluation trend: annotators generally rate StreamDial sessions as more informative and flexible than matched public-data dialogues, while still maintaining comparable coherence and naturalness.
We do not use this human evaluation to replace the reported DST results; instead, it serves as an independent validation channel for the intrinsic quality claims.
We will release the annotation guidelines together with the dataset documentation.

\subsection{Extrinsic Evaluation: Dialogue State Tracking}
\label{sec:extrinsic_eval}

We evaluate downstream utility on Dialogue State Tracking (DST) using Joint Goal Accuracy (JGA) and Slot-value F1.
Results are reported on three backbones (Qwen3-1.7B, Qwen3-8B, Gemma3-4B) under five training strategies: Zero-shot (No SFT), SFT w/ Public Data (Baseline), + Seed-Only, + Stream-Only, and + StreamDial (Hybrid).

A key protocol detail is that we keep the training size fixed at 2,000 dialogues for fair comparison.
The baseline model is trained on 2,000 public dialogues.
For each ``+'' setting, we replace half of the baseline data with synthesized data, i.e., 1,000 public dialogues + 1,000 synthesized dialogues from the corresponding source (Seed-Only, Stream-Only, or StreamDial-Hybrid).
Therefore, Table~\ref{tab:dst} measures the incremental utility of each synthesized source under a controlled 2k budget, rather than a change in training scale.

\vspace{0.5em}
\noindent\textbf{Main DST results.}
Table~\ref{tab:dst} reports public-test performance.
Compared with SFT w/ Public Data (Baseline), all three mixed settings improve DST across backbones, showing that synthesized dialogues provide complementary supervision beyond public corpora.
Among them, + StreamDial (Hybrid) gives the strongest overall results, indicating that combining seed scaffolding and stream-grounded signals is more effective than using either source alone.
Zero-shot performance is substantially lower, confirming that supervised adaptation remains necessary for reliable state tracking.

\vspace{0.5em}
\noindent\textbf{Cross-domain generalization on RiSAWOZ.}
To further test robustness, we evaluate Automotive, Restaurant, and Hotel separately.
For readability, the main text reports a representative strong backbone (Qwen3-8B), while full results for all backbones are provided in Appendix~\ref{app:cross_domain_full}.
As shown in Table~\ref{tab:cross_domain_qwen8b}, StreamDial-enhanced training achieves clear gains in Automotive and Restaurant, and remains competitive in Hotel, suggesting that the benefit of synthesized supervision generalizes across distinct service domains rather than being tied to a single scenario.

\begin{table}[h]
    \centering
    \small
    \caption{Cross-domain DST results on RiSAWOZ (Qwen3-8B, Public test set).}
    \label{tab:cross_domain_qwen8b}
    \setlength{\tabcolsep}{6pt}
    \renewcommand{\arraystretch}{1.12}
    \begin{tabular}{lcc}
        \toprule
        \textbf{Domain} & \textbf{Training Strategy} & \textbf{JGA(\%) / F1(\%)} \\
        \midrule
        \multirow{3}{*}{Automotive}
        & Zero-shot (No SFT)               & 54.04 / 73.40 \\
        & SFT w/ Public Data (Baseline)    & 93.94 / 98.56 \\
        & \hspace{0.8em}+ StreamDial (Hybrid) & \textbf{96.72 / 99.41} \\
        \midrule
        \multirow{3}{*}{Restaurant}
        & Zero-shot (No SFT)               & 56.18 / 81.40 \\
        & SFT w/ Public Data (Baseline)    & 90.44 / 97.91 \\
        & \hspace{0.8em}+ StreamDial (Hybrid) & \textbf{90.84 / 97.97} \\
        \midrule
        \multirow{3}{*}{Hotel}
        & Zero-shot (No SFT)               & 54.49 / 79.31 \\
        & SFT w/ Public Data (Baseline)    & \textbf{97.67 / 99.61} \\
        & \hspace{0.8em}+ StreamDial (Hybrid) & 96.01 / 99.36 \\
        \bottomrule
    \end{tabular}
\end{table}

To further analyze why synthesized data improves DST, we provide a slot-distribution study in Appendix~\ref{app:slot_distribution}.
The analysis compares slot coverage balance and slot-value diversity across data sources, and shows that different synthesis sources provide complementary supervision signals.

\subsection{Multilingual Transfer Evaluation}
\label{sec:cross_lingual}

We further evaluate multilingual transfer on X-RiSAWOZ in the Automotive domain, with English, French, and Korean as target languages.
To keep the comparison aligned with Section~\ref{sec:extrinsic_eval}, we use the same controlled training budget: 2,000 dialogues per run.
The baseline uses 2,000 target-language public dialogues.
Each ``+'' setting replaces half of the baseline set with synthesized data, i.e., 1,000 target-language public dialogues + 1,000 translated synthesized dialogues (Seed-Only, Stream-Only, or StreamDial-Hybrid), where the synthesized dialogues are translated from Chinese into the corresponding target language.
To reduce translation-induced variance, we use the same translation pipeline and post-processing rules across all target languages and synthesis settings.
This design isolates the effect of synthesized supervision under a fixed data scale, rather than conflating gains with larger training size.

For readability, the main text focuses on Qwen3-8B and compares SFT w/ Public Data (Baseline) against + StreamDial (Hybrid), while full multilingual results for all backbones and all settings are deferred to Appendix~\ref{app:xrisawoz_full}.
As shown in Table~\ref{tab:xrisawoz_qwen8b_main}, StreamDial improves both JGA and Slot-value F1 (reported in percentage points) in all three target languages.
The gains are larger in English and Korean, and smaller but still positive in French.

These Qwen3-8B results indicate that stream-grounded synthesis remains effective beyond the source language and can transfer useful dialogue structures to new languages under the same supervision budget.
At the same time, the smaller margin in French suggests that language-specific realization patterns can affect slot extraction stability after transfer.
Overall, the multilingual evidence supports the same conclusion as the monolingual setting: synthesized dialogues provide complementary supervision for DST when the training budget is controlled.

\begin{table}[h]
    \centering
    \small
    \caption{Cross-lingual DST on X-RiSAWOZ (Automotive), Qwen3-8B (\%).}
    \label{tab:xrisawoz_qwen8b_main}
    \setlength{\tabcolsep}{6pt}
    \renewcommand{\arraystretch}{1.12}
    \begin{tabular}{lcc}
        \toprule
        \textbf{Language} & \textbf{Training Strategy} & \textbf{JGA(\%) / F1(\%)} \\
        \midrule
        \multirow{3}{*}{English}
        & Zero-shot (No SFT)                  & 18.94 / 32.40 \\
        & SFT w/ Public Data (Baseline)       & 84.85 / 95.43 \\
        & \hspace{0.8em}+ StreamDial (Hybrid) & \textbf{87.63 / 96.86} \\
        \midrule
        \multirow{3}{*}{French}
        & Zero-shot (No SFT)                  & 16.41 / 39.32 \\
        & SFT w/ Public Data (Baseline)       & 80.30 / 94.23 \\
        & \hspace{0.8em}+ StreamDial (Hybrid) & \textbf{81.06 / 94.45} \\
        \midrule
        \multirow{3}{*}{Korean}
        & Zero-shot (No SFT)                  & 23.74 / 45.66 \\
        & SFT w/ Public Data (Baseline)       & 84.85 / 95.98 \\
        & \hspace{0.8em}+ StreamDial (Hybrid) & \textbf{85.61 / 96.03} \\
        \bottomrule
    \end{tabular}
\end{table}

\section{Limitations}
\label{sec:limitations}

Despite the promising results, this work has several limitations.

First, although streaming media provides rich and timely interaction signals, source quality is inherently noisy and uneven across platforms and creators.
We mitigate this issue with source filtering, domain lexicons, ASR correction, retrieval-based evidence checking, and consistency validation, but residual errors from ASR, colloquial ambiguity, and incomplete context may still propagate into downstream synthesis.

Second, StreamDial currently covers three service domains (Automotive, Restaurant, and Hotel). While these domains exhibit diverse service patterns, broader validation in additional verticals (e.g., healthcare consultation, education advising, legal assistance) is still needed to assess domain scalability.

Third, our evaluation emphasizes intrinsic dialogue quality, human validation, and downstream DST utility. Although these metrics are informative, they do not fully capture all aspects of real-world deployment, such as factual robustness under rapidly changing knowledge, user trust, and long-horizon task completion in interactive settings. The human-evaluation set used in this version covers 1,431 sessions, and we will release the annotation guidelines together with the dataset documentation.

Fourth, multilingual results are encouraging, but transfer quality can vary across languages due to differences in linguistic realization and slot expression patterns. Further work is needed on language-aware synthesis and calibration to reduce transfer variance. We are preparing translated dataset versions and associated translation metadata to reduce the reproducibility gap for cross-lingual experiments.

Finally, while the framework relies on publicly available media and structured processing, responsible data use remains important. Future releases should continue to strengthen documentation on data governance, filtering criteria, and ethical safeguards.

\section{Conclusion}
\label{sec:conclusion}

We presented Stream, a data-centric framework for synthesizing high-value task-oriented dialogues from publicly available streaming media. 
By coupling Streaming Signal Ingestion, Adaptive Persona Synthesis, Conversational Blueprint construction, and RAG-enhanced Interactive Dialogue Generation, Stream transforms noisy rich-media interactions into structured supervision signals for dialogue learning.

Based on this framework, we released StreamDial, a large-scale multi-domain dataset with 87,498 sessions and 1,497,320 turns, where each session is represented as a structured quadruplet $\langle P_u, P_a, B, H \rangle$. 
Empirical results show that StreamDial improves intrinsic dialogue quality and enhances downstream DST performance over strong baselines under controlled training budgets, with encouraging multilingual transfer on Qwen3-8B.

Overall, our findings suggest that streaming media is a practical and scalable data source for constructing strategy-rich task-oriented dialogue corpora. 
We hope Stream and StreamDial can serve as useful resources for future research on robust, knowledge-aware, and decision-oriented dialogue systems.

\bibliographystyle{ACM-Reference-Format}
\bibliography{acmart}

\clearpage

\appendix

\section{Chinese Version of the Case Study}
\label{app:case_study_zh}

\noindent\textbf{Note.}\quad
StreamDial is primarily collected and synthesized in Chinese.
For transparency, we provide the original Chinese case used in Figure~\ref{fig:case_study_streamdial}.
Different from the main paper (which shows an English translation for readability), this appendix keeps the dataset-native JSON structure (dialogue id, linked persona/plan ids, and turn-level records).

\vspace{0.6em}
\begin{figure*}[htb]
    \centering
    \includegraphics[width=\linewidth]{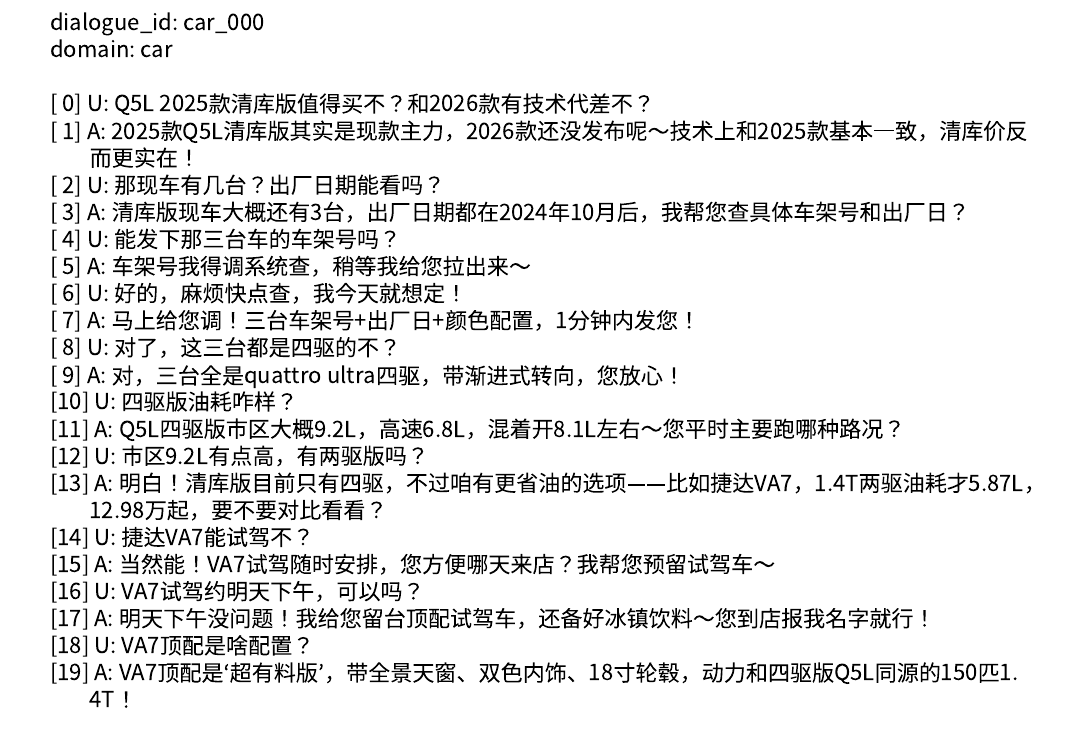}
    \caption{
    Dataset-format Chinese case study (Automotive), rendered from the original JSON entry.
    The example corresponds to the case shown in Figure~\ref{fig:case_study_streamdial}.
    }
    \label{fig:case_study_zh_json}
\end{figure*}

\clearpage
\section{Intrinsic Scores Underlying the Radar Plot}
\label{app:radar_scores}

Table~\ref{tab:radar_scores} reports the exact intrinsic scores used to generate Figure~\ref{fig:intrinsic_radar} in Section~\ref{sec:intrinsic_eval}.
We list per-judge scores for Open Data (Baseline) and \textbf{StreamDial (Hybrid)} over all six evaluation dimensions.

\begin{table*}[htb]
    \centering
    \small
    \setlength{\tabcolsep}{7pt}
    \caption{Per-judge intrinsic quality scores (1--10) used to plot Figure~\ref{fig:intrinsic_radar}.}
    \label{tab:radar_scores}
    \begin{tabular}{llcccccc}
        \toprule
        \textbf{Judge} & \textbf{Data} & \textbf{Coh} & \textbf{Info} & \textbf{Nat} & \textbf{Div} & \textbf{Flex} & \textbf{Overall} \\
        \midrule
        \multirow{2}{*}{Qwen3-Max}
        & Open Data (Baseline) & 8.06 & 6.04 & 7.52 & 4.68 & 5.68 & 6.32 \\
        & \textbf{StreamDial (Hybrid)} & 8.99 & 9.89 & 8.97 & 7.88 & 9.89 & 8.98 \\
        \midrule
        \multirow{2}{*}{GPT-5.2}
        & Open Data (Baseline) & 8.30 & 5.12 & 7.12 & 4.04 & 5.10 & 5.89 \\
        & \textbf{StreamDial (Hybrid)} & 8.67 & 7.93 & 7.63 & 6.57 & 8.44 & 7.81 \\
        \midrule
        \multirow{2}{*}{Gemini3-Pro}
        & Open Data (Baseline) & 8.59 & 6.84 & 7.64 & 5.52 & 6.11 & 6.91 \\
        & \textbf{StreamDial (Hybrid)} & 9.30 & 8.65 & 8.72 & 7.88 & 9.31 & 8.82 \\
        \bottomrule
    \end{tabular}
\end{table*}

\clearpage
\section{Additional Cross-domain DST Results}
\label{app:cross_domain_full}

This appendix reports the full cross-domain DST results on RiSAWOZ (Chinese), corresponding to Section~\ref{sec:extrinsic_eval}.  
In the main text, we present a representative backbone for readability; here we provide complete results for all reported backbones.

\begin{table*}[t]
    \centering
    \small
    \caption{Full cross-domain DST performance on RiSAWOZ (Public test set).}
    \label{tab:cross_domain_full}
    \setlength{\tabcolsep}{7pt}
    \renewcommand{\arraystretch}{1.15}
    \begin{tabular}{ll|cc|cc|cc}
        \toprule
        \textbf{Domain} & \textbf{Training Strategy}
        & \multicolumn{2}{c|}{\textbf{Qwen3-1.7B}}
        & \multicolumn{2}{c|}{\textbf{Qwen3-8B}}
        & \multicolumn{2}{c}{\textbf{Gemma3-4B}} \\
        & & \textbf{JGA(\%)} & \textbf{F1(\%)}
          & \textbf{JGA(\%)} & \textbf{F1(\%)}
          & \textbf{JGA(\%)} & \textbf{F1(\%)} \\
        \midrule
        \multirow{3}{*}{Automotive}
        & Zero-shot (No SFT) & 45.71 & 74.61 & 54.04 & 73.40 & 33.59 & 67.93 \\
        & SFT w/ Public Data (Baseline) & 93.43 & 98.81 & 93.94 & 98.56 & 95.71 & 98.78 \\
        & \hspace{0.8em}+ StreamDial (Hybrid) & \textbf{96.97} & \textbf{99.37} & \textbf{96.72} & \textbf{99.41} & \textbf{97.98} & \textbf{99.54} \\
        \midrule

        \multirow{3}{*}{Restaurant}
        & Zero-shot (No SFT) & 20.72 & 52.99 & 56.18 & 81.40 & 44.62 & 74.30 \\
        & SFT w/ Public Data (Baseline) & \textbf{90.44} & \textbf{97.64} & 90.44 & 97.91 & \textbf{92.03} & \textbf{97.97} \\
        & \hspace{0.8em}+ StreamDial (Hybrid) & 89.64 & 97.49 & \textbf{90.84} & \textbf{97.97} & 91.24 & 97.98 \\
        \midrule

        \multirow{3}{*}{Hotel}
        & Zero-shot (No SFT) & 14.62 & 59.37 & 54.49 & 79.31 & 36.54 & 73.11 \\
        & SFT w/ Public Data (Baseline) & 92.03 & 98.12 & \textbf{97.67} & \textbf{99.61} & 89.70 & 97.43 \\
        & \hspace{0.8em}+ StreamDial (Hybrid) & \textbf{92.36} & \textbf{98.22} & 96.01 & 99.36 & \textbf{96.35} & \textbf{99.26} \\
        \bottomrule
    \end{tabular}
\end{table*}

\clearpage
\section{Full Multilingual Results on X-RiSAWOZ}
\label{app:xrisawoz_full}

This appendix provides full cross-lingual DST results for X-RiSAWOZ (Automotive), including all reported backbones and training configurations.

\begin{table*}[t]
    \centering
    \small
    \caption{Full cross-lingual DST performance on X-RiSAWOZ (Automotive).}
    \label{tab:xrisawoz_full}
    \setlength{\tabcolsep}{5pt}
    \renewcommand{\arraystretch}{1.10}
    \begin{tabular}{ll|cc|cc|cc|cc}
        \toprule
        \textbf{Language} & \textbf{Training Strategy}
        & \multicolumn{2}{c|}{\textbf{Qwen3-1.7B}}
        & \multicolumn{2}{c|}{\textbf{Qwen3-8B}}
        & \multicolumn{2}{c|}{\textbf{Gemma3-1B}}
        & \multicolumn{2}{c}{\textbf{Gemma3-4B}} \\
        &  & \textbf{JGA (\%)} & \textbf{F1 (\%)}
          & \textbf{JGA (\%)} & \textbf{F1 (\%)}
          & \textbf{JGA (\%)} & \textbf{F1 (\%)}
          & \textbf{JGA (\%)} & \textbf{F1 (\%)} \\
        \midrule

        \multirow{5}{*}{English}
        & Zero-shot (No SFT)                  & 23.74 & 49.12 & 18.94 & 32.40 & 16.92 & 35.75 & 23.74 & 56.28 \\
        & SFT w/ Public Data (Baseline)       & 85.61 & 95.60 & 84.85 & 95.43 & 81.82 & 94.23 & 86.62 & 95.82 \\
        & \hspace{0.8em}* StreamDial (Hybrid) & 87.88 & 96.75 & 87.63 & 96.86 & 84.34 & 95.79 & 87.63 & 96.43 \\
        & \hspace{0.8em}** StreamDial (Hybrid) & 85.61 & 95.84 & 87.63 & 96.86 & 78.54 & 93.76 & 84.09 & 95.26 \\
        \midrule

        \multirow{5}{*}{French}
        & Zero-shot (No SFT)                  & 18.69 & 41.50 & 16.41 & 39.32 & 20.45 & 38.80 & 16.16 & 48.91 \\
        & SFT w/ Public Data (Baseline)       & 80.05 & 94.55 & 80.30 & 94.23 & 72.47 & 91.34 & 77.53 & 93.32 \\
        & \hspace{0.8em}* StreamDial (Hybrid) & 74.24 & 92.04 & 81.06 & 94.45 & 68.18 & 90.04 & 77.53 & 93.32 \\
        & \hspace{0.8em}** StreamDial (Hybrid) & 74.24 & 91.94 & 81.06 & 94.45 & 68.18 & 90.04 & 77.53 & 93.32 \\
        \midrule

        \multirow{5}{*}{Korean}
        & Zero-shot (No SFT)                  & 19.19 & 38.77 & 23.74 & 45.66 & 15.91 & 32.00 & 25.25 & 61.66 \\
        & SFT w/ Public Data (Baseline)       & 82.07 & 94.90 & 84.85 & 95.98 & 71.21 & 89.75 & 83.33 & 95.29 \\
        & \hspace{0.8em}* StreamDial (Hybrid) & 84.60 & 95.53 & 85.61 & 96.03 & 78.28 & 93.47 & 87.63 & 96.67 \\
        & \hspace{0.8em}** StreamDial (Hybrid) & 84.60 & 95.53 & 85.61 & 96.03 & 78.28 & 93.47 & 87.63 & 96.67 \\
        \bottomrule
    \end{tabular}
\end{table*}

\noindent\textbf{Training Strategies:} 
For all cross-lingual experiments, we explore two different approaches to multilingual dialogue synthesis using the StreamDial framework:
\begin{itemize}
    \item \textbf{*Source to Target Language*}: In this approach, we first translate source language seed dialogues into the target language before training, followed by generating target-language dialogues.
    \item \textbf{**Source Language to Target Language**}: In this strategy, we first train the model using the source language seed dialogues, generate corresponding source-language synthetic dialogues, and then translate them into the target language for training.
\end{itemize}

The results shown in the table indicate that the *Source to Target Language* approach leads to better performance in both English and Korean, while the **Source Language to Target Language** approach also yields competitive results, especially in French.

The table demonstrates the performance of models trained with StreamDial (Hybrid) under the two distinct training strategies. The *Source to Target Language* approach generally provides the best cross-lingual transfer, especially in the case of English and Korean. However, the results in French show a slightly less optimal transfer, possibly due to the inherent complexity of the French language structure affecting slot extraction accuracy.

\clearpage
\section{Slot Distribution Analysis by Data Source}
\label{app:slot_distribution}

To better understand the performance differences in Section~\ref{sec:extrinsic_eval}, we analyze slot statistics in the Automotive domain across five training sources:
\textbf{train} (original training split), \textbf{train\_llm\_slot (baseline)}, \textbf{our with seed dialogue}, \textbf{our with stream data}, and \textbf{our with hybrid data}.

We report two metrics for each slot type:
(1) \textbf{Coverage (\%)}: percentage of dialogues that contain this slot;
(2) \textbf{Values}: number of distinct values observed for this slot.
At the bottom of the table, we summarize (i) variance of slot coverage across slot types and (ii) average Values across slot types.

\begin{table*}[t]
    \centering
    \small
    \caption{Slot distribution statistics across data sources (Automotive).}
    \label{tab:slot_distribution_full}
    \setlength{\tabcolsep}{4.5pt}
    \renewcommand{\arraystretch}{1.1}
    \begin{tabular}{l|cc|cc|cc|cc|cc}
        \toprule
        \multirow{2}{*}{\textbf{Slot Type}}
        & \multicolumn{2}{c|}{\textbf{train}}
        & \multicolumn{2}{c|}{\textbf{train\_llm\_slot (baseline)}}
        & \multicolumn{2}{c|}{\textbf{our w/ seed dialogue}}
        & \multicolumn{2}{c|}{\textbf{our w/ stream data}}
        & \multicolumn{2}{c}{\textbf{our w/ hybrid data}} \\
        & Coverage & Values & Coverage & Values & Coverage & Values & Coverage & Values & Coverage & Values \\
        \midrule
        Power Level      & 5.13\%  & 2  & 0.32\%  & 1  & 10.69\% & 15 & 42.14\% & 29 & 26.42\% & 27 \\
        Manufacturer     & 19.71\% & 3  & 24.20\% & 5  & 40.25\% & 11 & 58.49\% & 11 & 57.86\% & 11 \\
        Seat Count       & 28.53\% & 3  & 36.22\% & 6  & 46.54\% & 7  & 4.40\%  & 2  & 27.67\% & 7  \\
        Price Range      & 39.90\% & 4  & 40.22\% & 6  & 98.11\% & 55 & 70.44\% & 71 & 88.68\% & 80 \\
        Fuel Consumption & 0.16\%  & 1  & 0.00\%  & 0  & 37.11\% & 36 & 3.14\%  & 5  & 11.32\% & 18 \\
        Vehicle Class    & 19.71\% & 4  & 20.35\% & 5  & 31.45\% & 8  & 5.03\%  & 3  & 29.56\% & 10 \\
        Energy Type      & 20.35\% & 2  & 20.51\% & 3  & 37.74\% & 5  & 8.81\%  & 4  & 23.27\% & 5  \\
        Body Type        & 79.17\% & 4  & 85.42\% & 6  & 88.05\% & 7  & 28.93\% & 6  & 79.25\% & 9  \\
        Series           & 94.07\% & 27 & 82.05\% & 34 & 86.79\% & 36 & 100.00\% & 76 & 92.45\% & 73 \\
        Drivetrain       & 21.31\% & 2  & 20.83\% & 2  & 47.17\% & 4  & 31.45\% & 4  & 38.99\% & 2  \\
        \midrule
        Coverage Variance (\(\downarrow\)) & \multicolumn{2}{c|}{8.43}
                                            & \multicolumn{2}{c|}{7.91}
                                            & \multicolumn{2}{c|}{\textbf{7.38}}
                                            & \multicolumn{2}{c|}{9.67}
                                            & \multicolumn{2}{c}{7.96} \\
        Avg. Values (\(\uparrow\))        & \multicolumn{2}{c|}{5.2}
                                            & \multicolumn{2}{c|}{6.8}
                                            & \multicolumn{2}{c|}{18.4}
                                            & \multicolumn{2}{c|}{21.1}
                                            & \multicolumn{2}{c}{\textbf{24.2}} \\
        \bottomrule
    \end{tabular}
\end{table*}

\noindent\textbf{Key observations.}
Compared with train\_llm\_slot (baseline), \textbf{our w/ seed dialogue} reduces coverage variance (7.91 \(\rightarrow\) 7.38), indicating a more balanced slot distribution across types.
At the same time, average slot-value diversity increases substantially (6.8 \(\rightarrow\) 18.4), suggesting richer semantic supervision.

\noindent
\textbf{our w/ stream data} yields the highest coverage variance (9.67), mainly because several slots are under-represented.
A likely reason is that those slots are mentioned less frequently in streaming interactions, which limits their exposure during persona and dialogue synthesis.

\noindent
For slot-value diversity, the ranking is:
\textbf{hybrid} (24.2) \(>\) \textbf{stream} (21.1) \(>\) \textbf{seed} (18.4).
This indicates that stream-grounded signals broaden value space, while hybrid composition further improves overall value richness through complementary sources.

\end{document}